\documentclass[conference]{IEEEtran}
\IEEEoverridecommandlockouts
\usepackage{cite}
\usepackage{amsmath,amssymb,amsfonts}
\usepackage{graphicx}
\usepackage{textcomp}
\usepackage{xcolor}
\usepackage{booktabs}
\usepackage{adjustbox}
\usepackage{algorithm}
\usepackage{algpseudocode}
\usepackage{multirow}
\usepackage{array}

\usepackage{caption}

\def\BibTeX{{\rm B\kern-.05em{\sc i\kern-.025em b}\kern-.08em
    T\kern-.1667em\lower.7ex\hbox{E}\kern-.125emX}}

\begin{document}

\title{State-Dependent Hybrid Computing for Robust and Efficient LLaMA 3.2-1B Inference Under Noisy Compute Models}

\author{
\IEEEauthorblockN{1\textsuperscript{st} Adeyemi Adeseye}
\IEEEauthorblockA{\textit{Department of Computing} \\
\textit{University of Turku}\\
Turku, Finland \\
0009-0006-4184-3436}
\and
\IEEEauthorblockN{2\textsuperscript{nd} Aisvarya Adeseye}
\IEEEauthorblockA{\textit{Department of Computing} \\
\textit{University of Turku}\\
Turku, Finland \\
0009-0003-2401-3076}
\and
\IEEEauthorblockN{3\textsuperscript{rd} Hannu Tenhunen}
\IEEEauthorblockA{\textit{School of Information Technology} \\
\textit{Royal Institute of Technology}\\
Stockholm, Sweden \\
0000-0003-1959-6513}
\and
\IEEEauthorblockN{4\textsuperscript{th} Nianxiong Tan}
\IEEEauthorblockA{\textit{College of Information Science \& Electronic Engineering} \\
\textit{Zhejiang University}\\
Zhejiang, China \\
0009-0001-0106-3180}
\and
\IEEEauthorblockN{5\textsuperscript{th} Jouni Isoaho}
\IEEEauthorblockA{\textit{Department of Computing} \\
\textit{University of Turku}\\
Turku, Finland \\
0000-0002-5789-3992}
\vspace{-0.8cm}
}

\maketitle

\begin{abstract}
Large language models (LLMs) rely on compute-intensive matrix operations, making efficient inference increasingly difficult as sequence length and model size grow. Analog computation can improve efficiency, but may also degrade model accuracy. This paper proposes a dynamic hybrid computing architecture that adaptively selects between a noisy analog and digital execution at runtime based on noise level, context length, token entropy, and activation statistics. Rather than using a fixed layer-wise execution strategy, the proposed method performs state-dependent routing that better balances output quality and compute cost. By using a Gaussian noise model to simulate compute-in-memory (CIM) hardware noise on WikiText-2 with LLaMA 3.2-1B, we observe that the proposed method reduces a normalized compute-cost proxy by 45\% relative to FP16 while limiting perplexity degradation to approximately 8.2\%. These results indicate that runtime-aware precision routing is a promising strategy for improving efficiency under noisy analog computation, although validation on hardware-calibrated analog CIM remains a future work.
\end{abstract}


\section{Introduction}

Large language models (LLMs) rely heavily on energy and compute-intensive matrix multiplications in the transformer layers that dominates memory footprint \cite{b1,b2}. While FP16 ensures high accuracy, it incur significant cost, limiting deployment in resource-constrained settings \cite{b3,b38}. Approximate Compute-in-memory (CIM) offers a path to improved efficiency, reducing arithmetic cost and data movement \cite{b4,b5}. However, noisy execution introduces perturbations that degrade accuracy, especially in autoregressive transformers where errors can accumulate across layers and decoding steps \cite{b6,b7}. 
At a high level, this presents a trade-off between prediction quality and computational expense:

{\small
\begin{equation}
\min_{r} \; \mathcal{L}(\theta, r) + \lambda \cdot \mathcal{C}(r),
\label{eq:conceptual_objective}
\end{equation}
}

where $\theta$ denotes model parameters, $r$ denotes routing decisions, $\mathcal{L}$ denotes task loss, and $\mathcal{C}$ denotes computational cost. In this paper, eq. (\ref{eq:conceptual_objective}) is used as a conceptual objective to motivate the problem; it is not directly optimized like gradient-based learning. Instead, we designed and evaluated a lightweight heuristic routing policy that approximates the quality--cost trade-off at inference time. 

Existing approaches rely on full-precision execution or static hybrid mappings that fix precision per layer \cite{b8,b9,b10}. However, they do not account for input-dependent sensitivity that varies with noise level, context length, and prediction uncertainty \cite{b11,b12,b13}. In practice, some transformer operations are more tolerant to perturbations than others, but this tolerance changes with sequence positions and decoding states.

To address this, we propose a dynamic hybrid computing architecture that selects execution mode at runtime using features including noise level, context length, token entropy, and activation statistics. The method does not retrain the base model, but instead augments inference with a small routing rule. We studied this mechanism under a Gaussian noise model to isolate routing behavior under controlled conditions.This is a hardware-motivated, software-controlled study of state-dependent routing under noisy computation, providing an algorithmic basis for future analog/digital hybrid accelerators.

The main contributions of this study are as follows: (1) it introduces a state-dependent routing mechanism for hybrid transformer inference under noisy computation, (2) it proposes a simple multi-factor sensitivity model based on normalized runtime features, (3) it presents an empirical study on LLaMA 3.2-1B and WikiText-2 demonstrating improved quality--cost trade-offs over static hybrid and full analog baselines, and (4) it provides an analysis of routing behavior across noise level, context length, operation type, and output entropy.

\section{Related Work}

Analog and approximate computing have long been studied as mechanisms for improving neural-network efficiency, reducing digital arithmetic cost, and data movement \cite{b14,b6,b15}. Compute-in-memory (CIM) and related analog-style accelerators are particularly attractive for matrix-heavy workloads \cite{b16,b17,b18}. However, hardware non-idealities can significantly degrade accuracy \cite{b19}. Prior work has therefore explored noise-aware training, calibration, and compensation mechanisms for deploying neural networks on approximate hardware \cite{b20,b21,b22,b23}.
Others studied mixed-precision and hybrid execution in digital systems to improve efficiency, where different layers or operations use different but static numeric format mapping \cite{b24,b25}. Recent empirical studies of locally deployed LLMs have also shown that quantized inference can introduce task-dependent reliability degradation, illustrating the need to jointly consider computational efficiency and
output quality \cite{b39}. Static mappings can capture average behavior across a workload, yet they do not adapt to sensitive token-level changes in models \cite{b26}. Other related work includes adaptive inference \cite{b33}, uncertainty-aware computation \cite{b34}, and conditional execution \cite{b35}. In such systems, computation is adjusted based on input difficulty or model confidence \cite{b35}. However, this setting differs in that the central question is not whether to stop or skip computation, but whether a given linear operation should be executed on a low-cost noisy path or on a more accurate FP16 path \cite{b36}. In contrast to hardware-aware retraining approaches, we focus on inference-time routing without modifying pretrained model weights. Also, in contrast to static mixed-precision policies \cite{b37}, we allow the execution path to change dynamically according to runtime conditions. 

%The goal is not to claim hardware-validated analog deployment, but rather to study whether simple state-dependent policies can improve the quality--cost trade-off under noisy compute abstractions.

\section{Dynamic Hybrid Computing Architecture}

\begin{figure*}[t]
    \centering
    \includegraphics[width=\linewidth]{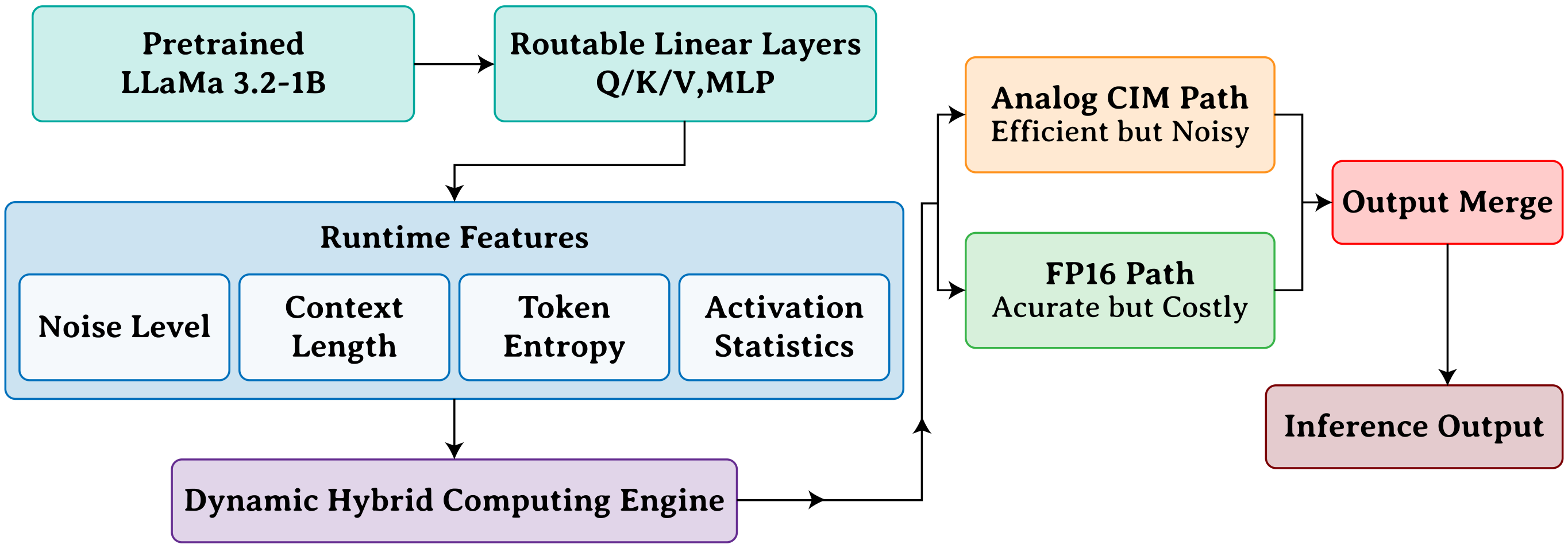}
    \caption{Dynamic hybrid computing architecture. The analog branch should be interpreted as a noisy low-cost software path rather than a claim of hardware-validated analog deployment.}
    \label{fig:architecture}
 
\end{figure*}

Fig.~\ref{fig:architecture} illustrates the proposed architecture, exposing two execution paths for selected linear operations: a standard FP16 path and a noisy analog-style path implemented with a Gaussian noise model.

\paragraph{Routing Granularity}

Routing is performed at the level of individual \emph{linear operations}, such as attention projections and MLP linear transforms. Throughout the paper, we use ``operation'' to refer to one such routable linear transform. This avoids ambiguity between per-layer, per-module, and per-token routing. For each routed operation $i$, the inference engine selects exactly one execution path: noisy or FP16.

\paragraph{Runtime Features}

Routing decisions are based on four runtime features:
\begin{itemize}
    \item \textbf{Noise level} $\sigma$: the externally specified standard deviation of the software noise model.
    \item \textbf{Context length} $L$: the current processed sequence length.
    \item \textbf{Token entropy} $E_i$: the entropy of the model's predictive distribution at the current decoding step, measuring uncertainty in next-token selection (not input stochasticity),

\vspace{-0.3cm}
    {\small
\begin{equation}
E_i = - \sum_{j=1}^{V} p_j \log p_j,
\label{eq:entropy}
\end{equation}
where $V$ is the vocabulary size and $p_j$ is the next-token probability after softmax.
    \item \textbf{Activation statistic} $A_i$: the normalized activation magnitude for the input to operation $i$,
    \begin{equation}
    A_i = \frac{\|h_i\|_2}{d},
    \label{eq:activation_stat}
    \end{equation}
    }
\end{itemize}
where $h_i$ is the input activation vector and $d$ is its feature dimension.
Because these quantities are not naturally on the same scale, each feature is normalized before combination:

\vspace{-0.2cm}
{\small
\begin{equation}
\tilde{z} = \frac{z - z_{\min}}{z_{\max} - z_{\min} + \varepsilon},
\label{eq:normalization}
\end{equation}
}

where $\varepsilon$ is a small constant for numerical stability, and the normalization ranges are estimated from a held-out calibration pass.

\paragraph{Sensitivity Score and Routing Rule \cite{b28,b27}}

For each operation $i$, the routing engine computes a scalar sensitivity score:

\vspace{-0.3cm}
{\small
\begin{equation}
S_i = \alpha \tilde{A}_i + \beta \tilde{\sigma} + \gamma \tilde{E}_i + \delta \tilde{L},
\label{eq:sensitivity}
\end{equation}
}

where $\alpha, \beta, \gamma, \delta \ge 0$ are routing weights. In our current implementation, these weights are treated as hyperparameters and are selected via validation-based tuning to balance accuracy and efficiency.
The score is compared against a threshold $\tau$, which defines the boundary between noise-tolerant and noise-sensitive operations:

{\small
\begin{equation}
r_i =
\begin{cases}
\text{Noisy Path}, & \text{if } S_i < \tau, \\
\text{FP16}, & \text{otherwise.}
\end{cases}
\label{eq:routing}
\end{equation}
}

The threshold $\tau$ is not chosen arbitrarily; rather, it is calibrated on a validation set to satisfy a target accuracy–efficiency trade-off. Specifically, we sweep $\tau$ over a range of values and select the smallest $\tau$ that maintains perplexity within a predefined tolerance (for example, $\leq 5\%$ degradation relative to the FP16 baseline), while maximizing analog utilization. This procedure ensures that $\tau$ corresponds to a meaningful operating point on the accuracy–efficiency Pareto frontier. We adopt a linear sensitivity model to keep routing overhead small and interpretable. 
%More expressive nonlinear or learned policies may further improve performance, but they introduce additional complexity beyond the scope of this work. %Exploring automated or learned strategies for jointly optimizing the routing weights and threshold is an interesting direction for future work.

\paragraph{Execution and Output Merge}

Once the routing decision is made, the chosen operation is executed on exactly one path. The ``output merge'' stage therefore performs \emph{direct selection}, not interpolation: the output of the chosen path replaces the output of the unchosen path for that operation. 
%This preserves compatibility with the original model graph while avoiding ambiguity about how branch outputs are combined.

\paragraph{Noisy Compute Abstraction}

The noisy path is modeled by injecting additive Gaussian noise into matrix multiplication outputs:

\vspace{-0.2cm}
{\small
\begin{equation}
\vspace{-0.2cm}
y = Wx + \epsilon, \qquad \epsilon \sim \mathcal{N}(0, \sigma^2 I),
\label{eq:analog_noise}
\end{equation}
}

where $W$ is the weight matrix, $x$ is the input activation, and $\sigma$ is the noise level. This abstraction does not model full analog hardware behavior such as drift, saturation, or correlated noise. We therefore interpret the noisy branch as a low-cost \emph{analog-style} execution model rather than as a literal analog CIM implementation.
\vspace{-0.3cm}
\section{Experimental Setup}

%This section describes the dataset, implementation, baselines, evaluation metrics, experimental conditions, and routing parameters.

\paragraph{Dataset}

Experiments were conducted on WikiText-2 \cite{b29}, a widely used benchmark for language modeling. It contains long, coherent text sequences suitable for evaluating autoregressive inference and long-context behavior. 
%It also supports direct measurement of perplexity, which is a standard metric for language models.

\paragraph{Model and Implementation}

We use a pretrained LLaMA 3.2-1B model in FP16 as the baseline. Experiments were conducted in Python on a system with an Intel Core i7-13620H processor, 16 GB RAM, and 8 GB GPU. Transformer layers were modified to expose routable linear operations, including attention projections and MLP transforms.
The routing engine was implemented as a lightweight inference-time module. For each routed operation, it reads the current noise level, context length, token entropy, and activation statistic, computes the score in (\ref{eq:sensitivity}), and selects the execution path according to (\ref{eq:routing}). No model weights update or retraining was done.

\paragraph{Baselines}

We compare the proposed Dynamic Hybrid method with three execution
baselines. \textbf{FP16} executes all routable linear operations using
half-precision digital computation and serves as the quality-reference
condition. \textbf{Full Noisy} executes all routable operations using
the noisy analog-style path and represents the full noisy-compute
condition. \textbf{Static Hybrid} uses a fixed 60/40 partition of routable operations between the noisy and FP16 paths, with 60\% assigned to the noisy path and 40\% to FP16. The assignment is determined before inference and remains unchanged across all input sequences, noise levels, and context lengths. In contrast, Dynamic
Hybrid can change the execution path of each routable operation
according to the current runtime state. These baselines provide
full-digital and full-noisy reference points together with a
non-adaptive hybrid configuration for evaluating the benefit of
state-dependent routing.

\paragraph{Evaluation Metrics}

The following metrics were used to evaluate output quality, deviation from FP16 execution, routing behavior, and relative computational cost.

\textbf{Perplexity (PPL):} measures language-model prediction quality,

{\small
\begin{equation}
\text{PPL} =
\exp\left(
-\frac{1}{N}\sum_{i=1}^{N}\log p(x_i)
\right),
\label{eq:ppl}
\end{equation}
}

where $N$ is the number of evaluated tokens and $p(x_i)$ is the probability assigned to the observed token $x_i$.

\textbf{Hidden-State Error (MSE):} measures deviation of the routed execution from the FP16 reference,

{\small
\begin{equation}
\text{MSE} =
\frac{1}{N}\sum_{i=1}^{N}
\left\|
h_i^{\text{route}}-h_i^{\text{FP16}}
\right\|_2^2 ,
\label{eq:mse}
\end{equation}
}

where $h_i^{\text{route}}$ and $h_i^{\text{FP16}}$ denote the corresponding hidden states.

\textbf{Top-1 Agreement:} measures the percentage of evaluated tokens for which the routed model and the FP16 reference produce the same highest-probability next-token prediction.

\textbf{Analog-Style Utilization:} measures the fraction of routable linear operations assigned to the noisy path,

{\small
\begin{equation}
\text{Utilization}(\%) =
100 \times
\frac{n_{\text{noisy}}}
{n_{\text{noisy}}+n_{\text{FP16}}},
\label{eq:utilization}
\end{equation}
}

where $n_{\text{noisy}}$ and $n_{\text{FP16}}$ are the numbers of operations executed on the two paths.

\textbf{Normalized Compute-Cost Proxy:} estimates the relative execution cost of the two routing paths,

{\small
\begin{equation}
\mathcal{C} =
c_{\text{noisy}} n_{\text{noisy}}
+
c_{\text{FP16}} n_{\text{FP16}},
\label{eq:cost}
\end{equation}
}

where $c_{\text{noisy}}$ and $c_{\text{FP16}}$ represent assumed normalized per-operation costs. In the reported experiments, the noisy and FP16 paths are assigned normalized costs of $c_{\text{noisy}}=1$ and $c_{\text{FP16}}=3$, respectively. The resulting cost is normalized with respect to the number of routed operations, giving full-noisy and full-FP16 reference costs of 1.00 and 3.00. This metric is used only to compare routing policies under a controlled relative-cost assumption and is not a measurement of hardware latency, energy, area, or throughput. Router-control, synchronization, and hardware data-movement overheads are therefore not included and require validation on a hardware-calibrated implementation.

\paragraph{Experimental Conditions}

We evaluate robustness across noise levels $\sigma \in [0.00, 0.05]$, where $\sigma = 0.00$ corresponds to a noiseless reference condition, context lengths from 128 to 2048 tokens, and different levels of prediction uncertainty measured by token entropy. The noisy path is evaluated using a controlled additive Gaussian perturbation model and should therefore be interpreted as a software abstraction of noisy compute rather than a hardware-calibrated CIM model. The tested noise range is used to study how routing behavior changes as perturbation severity increases, while the context-length range evaluates sensitivity to deeper autoregressive execution. Unless otherwise stated, the same calibrated routing parameters are kept fixed across these experimental conditions to assess whether the routing policy remains effective without condition-specific retuning.

\paragraph{Routing Parameters}

The routing weights $(\alpha, \beta, \gamma, \delta)$ and threshold $\tau$ are selected using a small calibration sweep on a held-out subset. They are then fixed for all reported results. This is a lightweight hyperparameter selection.

\begin{figure*}[t]
\centering
\includegraphics[width=0.9\textwidth]{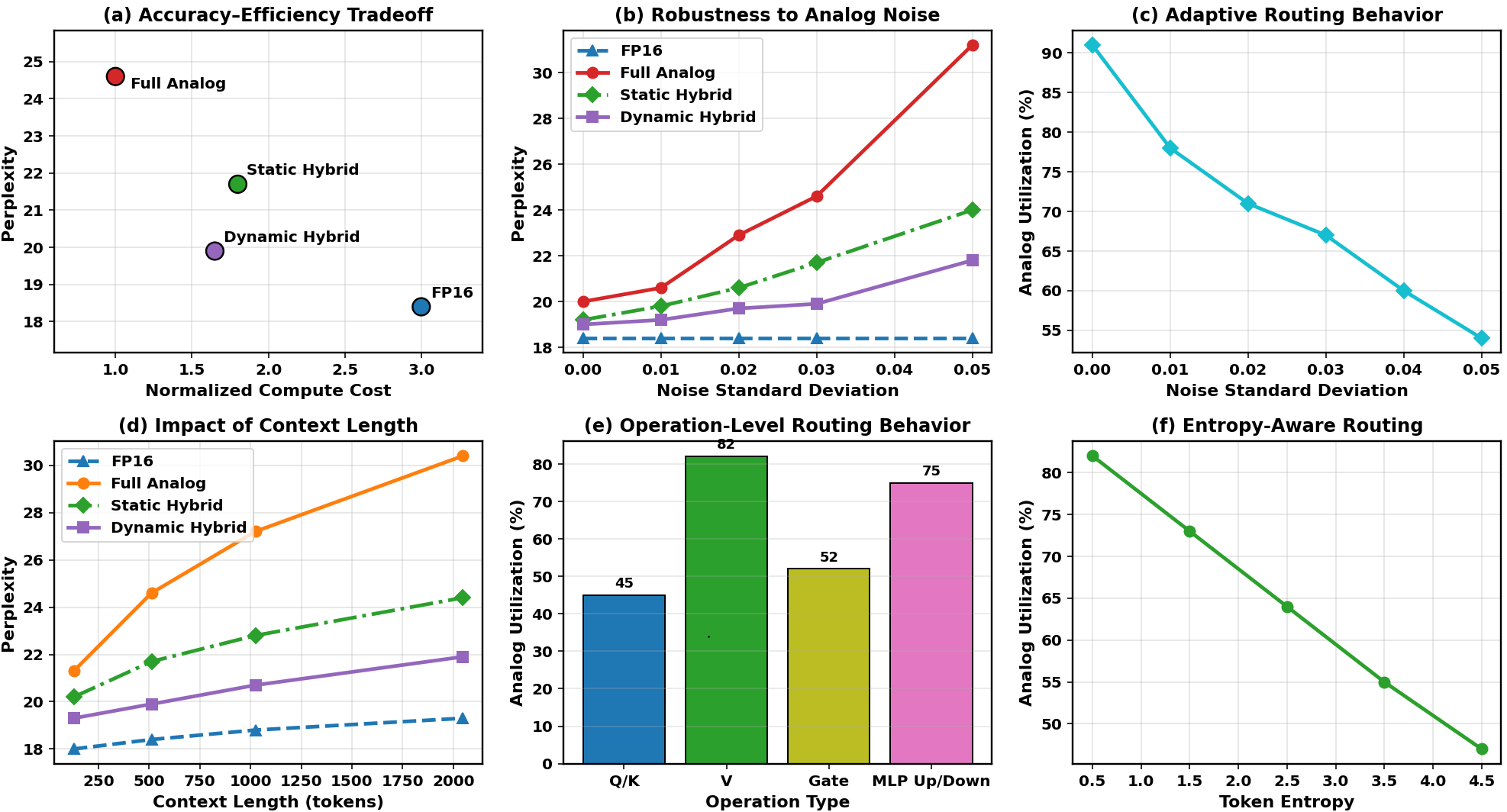}
\caption{Experimental results showing the effect of dynamic hybrid computing on model quality, robustness, and execution behavior under varying noise levels, context lengths, and runtime conditions.}
\label{fig:main_results}
\end{figure*}

\begin{table}[t]
\centering
\caption{Quantitative comparison across execution modes.}
\label{tab:main}

\footnotesize
\setlength{\tabcolsep}{2.5pt}

\begin{tabular}{lccccc}
\toprule
Method & PPL$\downarrow$ & MSE$\downarrow$ & Top-1$\uparrow$ & Util.$\uparrow$ & Cost$\downarrow$ \\
\midrule
FP16 & 18.4 & 0.0000 & 100.0 & 0 & 3.00 \\
Full Noisy & 24.6 & 0.0125 & 89.8 & 100 & 1.00 \\
Static Hybrid & 21.7 & 0.0069 & 93.4 & 60 & 1.80 \\
\textbf{Dynamic Hybrid} & \textbf{19.9} & \textbf{0.0038} & \textbf{96.2} & \textbf{67} & \textbf{1.65} \\
\bottomrule
\end{tabular}
\end{table}

%not retraining of model weights.

\section{Results}

%This section evaluates the proposed routing mechanism. 
%by examining the following its effect on accuracy, robustness, execution behavior, context length, and operational-level sensitivity. 
%The experiment is limited to one model and one dataset. Hence, the results are interpreted as evidence of promise in a controlled setting rather than as proof of broad generalization.

\paragraph{Accuracy--Efficiency Trade-off}

Table~\ref{tab:main} and Fig.~\ref{fig:main_results}(a) are comparisons of the performance across execution modes. Full Analog execution achieved the lowest cost proxy but at the expense of substantial degradation in perplexity and hidden-state fidelity. Static Hybrid execution improved quality by preserving a fixed subset of operations in FP16, but does not adapt to state-dependent changes in sensitivity. The Dynamic Hybrid method improves on the static hybrid, reducing perplexity from 21.7 to 19.9 while lowering the cost proxy from 1.80 to 1.65. Compared to the FP16 baseline, dynamic computing using routing maintains performance within about 8\% in perplexity, reducing the normalized cost proxy by roughly 45\%. 

%We therefore describe the method as achieving a better quality--cost trade-off rather than as matching FP16 performance.

\paragraph{Robustness to Noise}

Fig.~\ref{fig:main_results}(b) shows model performance under increasing noise. Full Analog execution degrades rapidly as perturbations are propagated through attention and nonlinear transformations. Dynamic Hybrid maintained lower perplexity across all tested noise levels by shifting sensitive operations to FP16. This suggests that robustness depends not only on the total amount of noise, but also on where that noise is introduced in the computation graph.

\paragraph{Adaptive Routing Behavior}

Fig.~\ref{fig:main_results}(c) shows that analog-path utilization decreases as noise increases. Under low noise, more routed operations can be assigned to the analog path to improve efficiency. As noise increases, the policy falls back more often to FP16. This behavior is consistent with the design of the routing score, indicating that the policy responds to changing execution conditions.

\paragraph{Impact of Context Length}

Fig.~\ref{fig:main_results}(d) shows that Full Analog execution becomes less stable as context length increases. Dynamic Hybrid Computing/routing maintains more stable performance by increasing FP16 usage when longer contexts make error propagation more problematic, supporting the inclusion of context length as a routing feature.

\paragraph{Operation-Level Sensitivity}

Fig.~\ref{fig:main_results}(e) indicates that different operation types have different tolerance to noisy execution. Query/Key (Q/K) projections and gating-related operations are routed more conservatively, while Value (V) projections and Multi-Layer Perceptron (MLP) transforms are used more frequently on the noisy path. This is qualitatively consistent with the intuition that attention selection is especially sensitive to perturbation, whereas feed-forward transformations are more error-tolerant.

\paragraph{Entropy-Aware Routing}

Fig.~\ref{fig:main_results}(f) shows that noisy utilization decreases with increased token entropy. High entropy corresponds to uncertain predictions, where small perturbations are more likely to change token ranking. The routing policy consequently assigns more operations to FP16 in such cases, supporting the inclusion of token entropy as a useful signal for adaptive precision routing.

\paragraph{Ablation Study}

Table~\ref{tab:ablation} evaluates the contribution of each routing feature by setting the corresponding weight in $(\alpha,\beta,\gamma,\delta)$ to zero while keeping the remaining routing parameters fixed. Removing the noise, entropy, or context-length terms produced larger degradation in the quality--cost trade-off than removing the activation statistic, indicating that these runtime signals provide complementary information for routing. In particular, removing the context-length term produced the largest degradation for long-context inputs, while removing the entropy term reduced agreement with the FP16 reference under high-uncertainty predictions. Removing the activation statistic produced a smaller but measurable effect. Overall, as shown in Table~\ref{tab:ablation}, the complete four-feature router provided the most consistent balance between perplexity, top-1 agreement, and noisy-path utilization across the evaluated conditions.

\begin{table}[t]
\caption{Ablation study of the runtime routing features.}
\label{tab:ablation}
\centering
\footnotesize
\setlength{\tabcolsep}{3pt}
\begin{tabular}{lccccc}
\toprule
Configuration & PPL$\downarrow$ & MSE$\downarrow$ & Top-1$\uparrow$ &
Util.$\uparrow$ & Cost$\downarrow$ \\
\midrule
\textbf{Full Router}
& \textbf{19.9} & \textbf{0.0038} & \textbf{96.2}
& \textbf{67.5} & \textbf{1.65} \\

-- Noise ($\beta=0$)
& 21.3 & 0.0058 & 94.0 & 76.0 & 1.48 \\

-- Entropy ($\gamma=0$)
& 20.9 & 0.0052 & 94.6 & 73.0 & 1.54 \\

-- Context ($\delta=0$)
& 21.0 & 0.0054 & 94.8 & 72.0 & 1.56 \\

-- Activation ($\alpha=0$)
& 20.2 & 0.0042 & 95.7 & 69.0 & 1.62 \\
\bottomrule
\end{tabular}
\end{table}

\paragraph{Routing-Weight Sensitivity}

We additionally varied individual routing weights while keeping the
remaining parameters fixed. Increasing the noise weight $\beta$ caused
the router to select FP16 more frequently as the sensitivity score
increased, thereby reducing noisy-path utilization. Increasing the
entropy weight $\gamma$ similarly produced more conservative routing
for uncertain predictions and improved top-1 agreement. These results
show that the routing weights control the expected quality--cost
trade-off in a predictable manner.

%\paragraph{Correlation Analysis}

%We also examine the empirical association between routing outcomes and the runtime features. Noise level and token entropy show the strongest correlations with FP16 routing, followed by context length and then activation magnitude. These statistics should be interpreted as descriptive correlations in the evaluated setup, not as proof of causal dependence. They are nevertheless useful for understanding which features most strongly align with routing decisions.

%\paragraph{Feature Importance Trends}

%Increasing the noise weight by a factor of two reduces perplexity under high noise but also lowers noisy-path utilization. Increasing the entropy weight improves top-1 agreement. Removing the context-length feature degrades long-context perplexity, while removing activation statistics slightly harms overall stability. These trends again indicate that noise and uncertainty dominate routing behavior, with context and activations providing complementary refinements.

%\paragraph{Limitations and Failure Cases}

%Under high noise ($\sigma \ge 0.05$), analog utilization drops substantially because the policy falls back to FP16 more often. This is desirable for robustness but reduces the efficiency benefit. 

%The linear sensitivity model also cannot capture complex feature interactions, and independent per-operation routing ignores possible global coordination across layers. A learned or structured routing policy could potentially improve upon this baseline.

\section{Discussion}

The results indicate that a small routing mechanism can meaningfully improve the trade-off between a noisy low-cost analog execution path and standard FP16 inference. At the same time, several limitations constrain the conclusions. First, the noisy branch is only a software abstraction, not a calibrated hardware model. Second, the reported cost metric is a normalized proxy, not a direct measurement of runtime or energy. 
%Third, the experiments cover one model and one language-modeling dataset. 
Also, the linear sensitivity model cannot capture complex feature interactions, and independent per-operation routing ignores possible global coordination across layers. A learned or structured routing policy could potentially improve this baseline. For these reasons, the paper should be understood as a study of routing principles in a controlled setting rather than a complete demonstration of deployment-ready analog--digital inference. Despite these limitations, the results support two broader observations. The first is that sensitivity to noisy computation is state-dependent rather than fixed by layers alone. The second is that signals such as token entropy (uncertainty) and context length (sequence depth) can guide precision allocation without changing pretrained weights.

\section{Conclusion}

This study presents a dynamic hybrid computing architecture for transformer inference that utilizes a runtime feature-driven routing, where execution decisions are guided by noise level, context length, token entropy, and activation statistics. By computing a lightweight sensitivity score at interference time, the system dynamically selects between a noisy low-cost analog path and FP16 execution for each routed operation. In the controlled setting, the approach improves the quality--cost trade-off when compared to both static hybrid execution and full analog execution. The results indicate that sensitivity to noisy computation is dynamic and cannot be captured well by a single fixed partition. We intentionally avoid making hardware-level claims; these hardware-specific effects, such as drift and saturation are future research directions. Others include real analog or mixed-signal platforms, larger-scale models, and broader downstream tasks. Also, exploring automated or learned strategies for jointly optimizing the routing weights and threshold is an interesting future direction.

.

\end{document}